\documentclass[conference]{IEEEtran}

\usepackage{hyperref}
\usepackage[sort&compress,numbers]{natbib}

\ifCLASSINFOpdf
  \usepackage[pdftex]{graphicx}
  \DeclareGraphicsExtensions{.pdf,.jpeg,.png}
\else
\fi

\usepackage{amsmath}
\usepackage{amsfonts}
\usepackage{amsthm}
\usepackage{newtxmath}  %

\DeclareMathOperator*{\argmax}{arg\,max}
\newcommand{\R}[0]{\mathbb{R}}

\newcommand*{\T}{^{\mkern-1.5mu\mathsf{T}}}

\usepackage{balance}

\begin{document}
\title{Control-Theoretic Analysis of \\ Shared Control Systems}

\author{\IEEEauthorblockN{Reuben M. Aronson}
\IEEEauthorblockA{Department of Computer Science\\
Tufts University\\
Medford, Massachusetts, USA\\
reuben.aronson@tufts.edu}
\and
\IEEEauthorblockN{Elaine Schaertl Short}
\IEEEauthorblockA{Department of Computer Science\\
Tufts University\\
Medford, Massachusetts, USA\\
elaine.short@tufts.edu}}

\maketitle

\begin{abstract}
Users of shared control systems change their behavior in the presence of assistance, which conflicts with assumpts about user behavior that some assistance methods make. In this paper, we propose an analysis technique to evaluate the user's experience with the assistive systems that bypasses required assumptions: we model the assistance as a dynamical system that can be analyzed using control theory techniques. We analyze the shared autonomy assistance algorithm and make several observations: we identify a problem with runaway goal confidence and propose a system adjustment to mitigate it, we demonstrate that the system inherently limits the possible actions available to the user, and we show that in a simplified setting, the effect of the assistance is to drive the system to the convex hull of the goals and, once there, add a layer of indirection between the user control and the system behavior. We conclude by discussing the possible uses of this analysis for the field.
\end{abstract}

\IEEEpeerreviewmaketitle

\section{Introduction}

Recent work~\cite{Aronson2024IntentionalAssistance} shows that users actively explore the behavior of assistive systems, reason about how they behave, and change their own inputs to drive the systems to their desired goals. The behavior is difficult to reconcile with assistance methods that require assumptions about user behavior, such as goal-predictive assistance systems~\cite{Losey2018AInteraction}. In fact, we theorize the problem may be more serious: if users always have the ``last word,'' analytical tools that require assumptions about user behavior or response will encounter recursive issues as users continually adapt to changes in the system.

In this paper, we propose an alternative analysis technique based on this insight: rather than model the user directly, we model the rest of the robot and assistance systems as a dynamical system that the user is controlling. This technique is not a substitute for other methods of developing assistance. Rather, it enables examining the user's experience independently from whatever assumptions about the user, environment, or task that the assistance relies on. In this way, all assistance can be viewed as changes to a dynamical system that the user is operating, and any adaptation or learning performed by the system becomes added dynamics for the robot.

To develop this approach, we turn to dynamical systems and control theory. These fields, which are commonly used in robotic systems, study how a system behaves over time in response to its inputs, internal dynamics, and outputs. Control theory further explores how to design rules for providing inputs to guarantee certain behaviors of the system while knowing only the outputs. Our key insight is that when operating a system with assistance, the \emph{user} is performing precisely this control operation however the assistance is designed. The assistance system defines the system \emph{dynamics} that the user controls, and designing assistance behavior is equivalent to designing the plant. Thus, we can use insights from control theory about what makes dynamics easy or hard to control and apply them to evaluating the assistance itself.

In this paper, we develop this control-theoretic analysis of shared control systems by analyzing of shared autonomy~\cite{javdani18}, a popular shared control formulation. We show that the assistance acts as a pure integrator, which has some undesirable properties, and use controls analogies to propose mitigations. We next show that the assistance restricts the possible actions that the system can take no matter the user's intentions. Finally, we discuss a simplified setting, free-space navigation, and show that the dynamics of the system reduce to the user moving a set point that the system settles at. We conclude with a discussion of next steps for this analytic approach and implications for the design of shared control systems.

\section{Background}
\subsection{Shared Control}

In a shared control framework, both the user and an assistive algorithm simultaneously control a robot~\cite{Cimolino2022TwoControl}. The assistance modifies the robot's behavior to drive the system in a way that the user ``actually means'' --- motivations for shared control include challenges of low-dimensional interfaces~\cite{muelling2017,Gopinath2021CustomizedRobots} or safety-critical situations~\cite{Okamura2004MethodsSurgery}. Goal-predictive assistance~\cite{Losey2018AInteraction} breaks the problem down into \emph{intent detection}, during which the system estimates the ``true'' goals of the user using various methods\cite{Jain2019ProbabilisticRobotics,Baker2007GoalPlanning,Ziebart2008MaximumLearning}, and \emph{arbitration}, during which the user's initiative is mixed with an assistive action generated based on the detected intent. These systems improve performance in many settings~\cite{Huntemann2008OnlineSteering,Dragan2013AControl,reddy2018rss,JunJeon2020SharedActions,Aronson2022GazeTeleoperation}. However, they require strong assumptions about user behavior which do not necessarily hold in practice~\cite{Aronson2024IntentionalAssistance}, thus benefiting from our approach. 

Other assistance methods focus directly on the dynamics. Most directly, virtual fixtures~\cite{Marayong2003SpatialFixtures} apply artificial forces to the robot's input response. More sophisticated approaches change the behavior of the robot to encourages successful task performance~\cite{Broad2017LearningDynamics,Yoneda2023ToAutonomy}. Our analysis broadens this assistance strategy to characterize assistance behavior that is not designed this way explicitly.

Though shared control research generally focuses on algorithmic advances, several papers consider the human experience of controlling a robot in a shared control paradigm~\cite{Rea2022StillInterfaces,Aronson2024IntentionalAssistance}. This work includes explicitly modeling how the user will adapt to the system~\cite{Nikolaidis2017Human-robotExperiments,Parekh2022RILI:Intent}, changing the system to make the user's behavior more informative~\cite{Gopinath2017ModeDisambiguation,Jonnavittula2022CommunicatingAutonomy}, and adjusting the user model based on observed behavior~\cite{JunJeon2020SharedActions,Zurek2021SituationalAutonomy,Newman2024BootstrappingCollaboration}.

\subsection{Control Theory}

Control theory analyzes how to drive a dynamical system to a particular desired output. Dynamical systems models consist of three components: the \emph{input} $u$, the \emph{state} $x$, and the \emph{output} $y$. Then, control theory answers the following question: Assume we can provide inputs $u$ to the system and observe output $y$, but we do not have direct access to $x$. How do we select the input $u$ to drive the system to produce some desired output $y^*$? For this paper, we examine assistance from the perspective of the user, who can provide input $u$ to the assistance system and observe output $y$ but does not have direct access to the internal state $x$. We then use insights from control theory about what makes dynamics easy or hard to control and apply them to designing the assistance itself.

Control theory techniques are common in robotics in general. Most low-level motor controllers are PID (proportional-integral-derivative) systems, while model-predictive control is used for short-horizon optimization of robot low-level commands. Here, we consider the inverse problem: how do we design the dynamics of a system so that it is easy for a controller to be designed that drives the system to an intended goal?

\section{Dynamical Analysis of Shared Autonomy}
To explore this assistance analysis, we present a worked example. We analyze the shared autonomy control paradigm as a dynamical system that the user controls and show that it acts as an integrator, which is known to cause unbounded lag when trying to change intended goals. We also show that the presence of the assistance fundamentally limits the actions that the user can take regardless of their input. Further, we show that in an obstacle-free navigation task, the system dynamics are fully determined by a set point (the expected value of the goal locations over the goal probabilities): the user adjusts the set point location by changing the goal probabilities, and the assistance drives the system to that set point.

\subsection{Background: Shared Autonomy}
We begin by briefly summarizing the shared autonomy algorithm presented in \citet{Javdani2015SharedOptimization}. The algorithm expects a pre-specified (finite) list of $k$ goals $G$ along with optimal policies for each goal defined by action-value functions $Q_g(x,a)$. It receives an input $u_t$ from the user at each time step, uses a Bayesian update framework to estimate the user's true goal within $G$ as a probability distribution, then takes an action that is optimal in expectation over the probability of the user's true goal. This assistive action is then applied to the system itself. The algorithm is intended to improve performance of the system for a user who knows their goals but cannot optimally drive the robot to achieve them; the robot uses its own (presumably) optimal policy to reach them instead. This system decreases task completion time and user effort required for reaching the goal.

At each time step, the system estimates the likelihood of an observed user action $u_t$ given each goal. Several models for the likelihood are used~\cite{Baker2007GoalPlanning,Ziebart2008MaximumLearning}; here, we will show the Bolzmann rationality formulation, which sets
\begin{equation}
    \label{eqn:likelihood}
    p(u_t|x_t,g) \propto \exp \beta Q_g(x_t, u_t)
\end{equation}
with $x_t$ the current state, $\beta$ a ``rationality'' parameter that varies from $\beta = 0$, indicating random user actions, to $\beta = \infty$, indicating perfect user actions. Note that this probability is conditioned on the stat $x_t$; this approach treats the user's input as informative separately from the actual robot state in which it was provided. For ease of notation, we drop $x_t$ throughout.

Next, this likelihood is used to update the assistance system's estimate of the user's goal, stored as a probability distribution $p(g|u_0, \cdots, u_t)$. Via Bayes' rule, we can set
\begin{equation}
    \label{eqn:bayes-update}
    p(g|u_0, \cdots, u_t) = \frac{1}{Z_t} p(u_t| g) p(g|u_0, \cdots, u_{t-1}),    
\end{equation}
with $Z_t = \sum_g p(g|u_0, \cdots, u_t)$ a normalization factor.

Finally, this system generates assistance by selecting an action $a_t$ that has the largest expected increase in value over the goal distribution, given by
\begin{equation}
    \label{eqn:assist-action}
    a_t = \argmax_{a'} \sum_g p(g|u_0, \cdots, u_t) Q_g(x_t, u_t).
\end{equation}
This equation is derived from applying the QMDP assumption to solving a POMDP representation of the user's behavior; see the paper for further details.

\subsection{Dynamical Systems Representation}
We now show how the same assistance system appears to the user as a dynamical system. To do so, we define the internal state of the system and its dynamics, how the state changes based on the input, and how the output of the system is derived from the state. 

First, for convenience, we define $\vec{Q}_t(u) = [ Q_{g_0}(u), Q_{g_1}(u), \cdots, Q_{g_{k-1}}(u) ]\T$ as a vector-valued function representing the value of taking the same action $u$ in the state $x_t$ with respect to each of the goals in turn. This transformation turns a possible action $u$ into a vector of length $k$ that describes how well $u$ aligns with each goal in turn.

Next, we define the state of the assistance system $\ell_t$ as
\begin{equation}
    \ell_t = [ \log p(g_0 | u_0, \cdots, u_t ), \cdots, \log p(g_{k-1} | u_0, \cdots, u_t ) ],
\end{equation}
the log of the goal probability vector used in each step. By taking the log of Eqn.~\ref{eqn:bayes-update} and substituting $\ell_t$ as appropriate, we obtain the system dynamics
\begin{equation}
    \ell_t = \ell_{t-1} + \log p(u_t|g) - \log Z_t.
\end{equation}

Note that this is a linear system in $\ell$, but is nonlinear in the user input $u_t$. However, the user contribution to the update is in a single additive term, so we can perform a change in variables to remove it. We define $v_t(g) = \log p(u_t|g)$ as contribution of the user's input towards the system's selection of goal $g$; using Eqn~\ref{eqn:likelihood}, we compute
\begin{equation}
    v_t(g) = \beta Q_g(u_t) - \log \sum_{u' \in U} \exp(\beta Q_g(u')).
\end{equation}
Using this equation, we can express the dynamics in terms of the input vector $\vec{v}_t = [v_t(g_0), v_t(g_1), \cdots, v_t(g_{k-1}]$ 
as 
\begin{equation}
    \ell_t = \ell_{t-1} + \vec{v}_t - \log Z_t.
\end{equation}

Next, we consider the output of the system $a_t$. Using Eqn.~\ref{eqn:assist-action} and substituting $\ell_t$, we find
\begin{equation}
    a_t = \argmax_{a'} \exp \ell_t \cdot \vec{Q}(a').
\end{equation}
Now, note that the term $Z_t$ is positive, uniform over $g$, and only used inside the $\argmax$ term, so it can be dropped entirely (though normalization may be required for numerical stability). Thus, our final system is
\begin{align*}
    \ell_t &= \ell_{t-1} + \vec{v}_t \\
    a_t &= \argmax_{a'} \exp \ell_t \cdot \vec{Q}_t(a').
\end{align*}

\subsection{Dynamics Analysis}
The system dynamics for shared autonomy are relatively simple. Though (generally) nonlinear functions $v_t(g) = v(x, u, g)$ are required to process the raw user input $u_t$ and compute the output action $a_t$, the system dynamics consists of a single integrator.

This representation gives us several insights into the behavior of the assistance system from the perspective of the user. Pure integrator systems are known to have infinite DC gain: if supplied a constant input, the system will increase without bound. In this case, providing a constant user command will cause the probabilities assigned to goals other than the most likely to decrease without bound. If the user has been pursuing one goal for some time and then switches to a different goal, the user must drive the system towards the new goal with an amount of effort proportional to what was spent achieving the first goal before the system will register a change in their intention. Though the original formulation of shared autonomy assumes that users will not change their goals, we can still analyze the system behavior if they do. 

Dynamical systems theory also gives us insight into how to mitigate this effect. Since we have full control over the dynamics of the goal update, we can introduce a new term to Eqn.~\ref{eqn:bayes-update} that adds internal stability to the system:
\begin{equation}
    \ell_t = k \ell_{t-1} + v_t,
\end{equation}
where $0 \leq k \leq 1$ is a stabilizing term that limits how extreme the values of $\ell_t$ can become. Given a consistent user input, the probability distribution will eventually reach a steady-state value (ignoring the effect of renormalization), so the effort to change the system to select a different goal remains bounded. Control theory offers other options as well, such as adding a term dependent on $\vec{v}_t - \vec{v}_{t-1}$ which monitors the \emph{change} in user input. The analysis suggests changes in the dynamics of the system that the user is operating which can be evaluated in future user studies. 

\begin{figure}
    \centering
    \includegraphics[width=0.8\columnwidth]{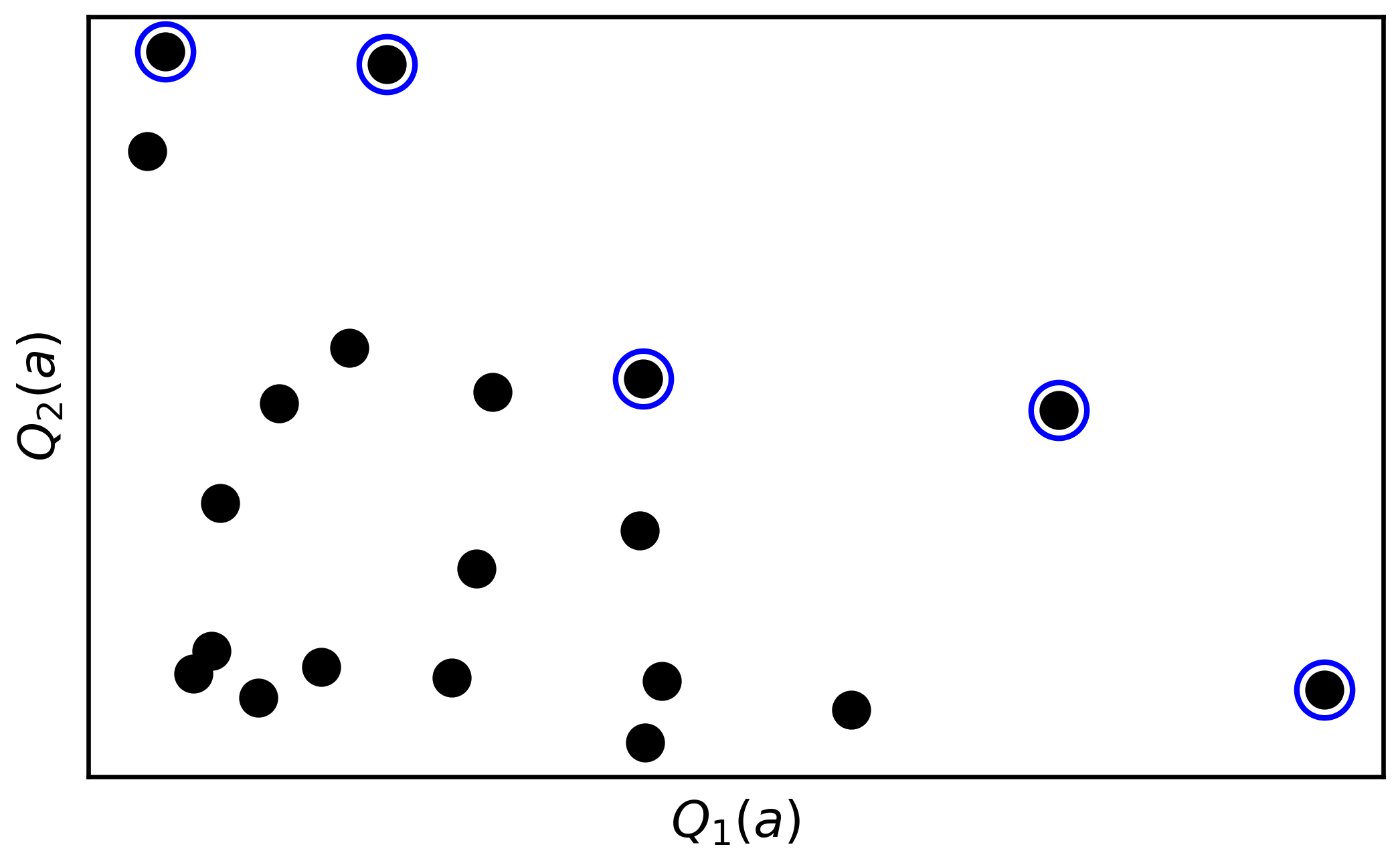}
    \caption{While there may be many possible actions that the system can take in some state $x$, here plotted with their Q-values corresponding to two different goals, the assistance system only ever selects actions that lie on the Pareto frontier, indicated by the blue circles.}
    \label{fig:pareto}
\end{figure}

\subsection{Experience of User Control}
Now, we turn to a different question: what are the possible actions that an operator can cause the system to produce? This question makes no sense in a robot-centric perspective: the goal-specific policy defines the optimal actions for each goal, so no additional actions are needed. %
However, users may want to drive the system to produce other actions based on their individual information or preferences, and we can use similar analysis tools to determine their options. To examine the possible actions, we consider Eqn.~\ref{eqn:assist-action}. For the system to select an action $a$, the action must be the $\argmax$ for \emph{some} value of $\ell_t$. Furthermore, since $\exp \ell_t > 0$, the term inside the $\argmax$ is a positive combination of $\vec{Q}$; by definition, then, $\vec{Q}_t(a)$ must lie on the Pareto frontier of $\{ \vec{Q}_t(a) : a \in A \}$. Therefore, not all actions $A$ are possible. In fact, the possible actions are a function only of the state $x_t$. User input can only select among the Pareto-optimal actions, and other actions are impossible to achieve no matter the user input (see Fig.~\ref{fig:pareto}).

With simplifying assumptions, we can go further. Consider a free-space navigation system with real states $S = \R^n$ and real unit actions $A = \{a \in \R^n : |a| = 1 \} \cup \{0\}$; we will denote these states and action as $\vec{x}$ and $\vec{a}$ for clarity. Further, let each goal $g \in S$ indicate a target state, which we will denote $\vec{g}$; $G$ represents a matrix $[g_0, g_1, \cdots, g_{k-1}]$ of all goals with each row corresponding to a goal. We posit a closed-form value function $Q_g(\vec{x},\vec{a}) = (\vec{g}-\vec{x})\T \vec{a}$ which computes how well the proposed action $\vec{a}$ aligns with the direction to the goal $(\vec{g}-\vec{x})$; with this assumption, the system is a discrete-time analogue of potential function navigation\footnote{On the domain $|g-x|>1$, with $V(x) = |g-x|$, $r(x) = (g-x)\T a - |g-x-a|$, and $T(x,a) = x + a$, this $Q$ satisfies the Bellman equation.}. Then, $\vec{Q}(a) = Q\T \vec{a}$ is in fact a matrix multiplication with 
\begin{align}
    Q = \begin{bmatrix}
    | & | & & | \\
    \vec{g}_0 - \vec{x} & \vec{g}_1 - \vec{x} & \cdots & \vec{g}_{k-1} - \vec{x} \\
    | & | & & | \\
    \end{bmatrix}.
\end{align}
In this scenario, we can solve the $\argmax$ term directly:
\begin{align*}
    \vec{a}^* &= \argmax_{\vec{a}} \exp \ell \cdot \vec{Q}(\vec{a})  \\
    &= \argmax_{\vec{a}} (\exp \ell)\T Q\T\vec{a} \\
    &=  \frac{Q \exp \ell}{|Q \exp \ell|} .
\end{align*}
Thus all assistance actions $\vec{a}$ are normalized positive combinations of the columns of $Q$, i.e., within the positive span of the directions towards each goal. 

From this analysis, we can draw several conclusions. First, the assistance can produce any action in $A$ precisely when the positive span of $Q$ includes the whole space. This scenario occurs when the origin lies within the convex hull of $Q$ or, equivalently, when the state $\vec{x}$ is inside the convex hull of the goals $\vec{g}$. Thus, the assistance dynamics have two phases: when the state is outside the convex hull of the goals, the assistance will drive the state towards the convex hull no matter the user input. The user can change the direction of approach among the available goals but cannot drive the system away from the convex hull. Once the system is within the convex hull, any action can be achieved by manipulating the assistance state $\ell_t$ so long as the system state remains within the convex hull; the user cannot drive the system outside the hull. This hull always exists because $|G|$ is finite (by assumption), though it can be degenerate; in this case, the system is driven to and remains in the hyperplane defined by the goals and by the convex hull within that hyperplane.

Another result of this analysis is that the system can be made to stop at any point within the convex hull of the goals. Formally, for all states $\vec{x} \in \text{conv}(\vec{g})$, there is some vector $v$ with all nonnegative elements such that $Q\,v = 0$. Since $v$ is nonnegative, we can select $\exp \ell = v$; for this probability state, the optimal action $a = \text{norm}(Q \exp \ell) = 0$. Conversely, all goal prediction states have a unique stable point at $\vec{g} \exp \ell = \sum_k p(k)g_k = x^*$, the convex combination of the goal locations with weights equal to the goal probabilities; this fact follows from solving $Q \exp \ell = 0$ for $\vec{x}$.

Putting everything together, shared autonomy imposes a two-phase system. When the system state is outside the acceptable region, defined by the (possibly lower-dimensional) convex hull of the goals, the system will drive it towards the convex hull independent of user input. Within the convex hull, though, the system supports any desired action, but through a layer of indirection. The user input updates the set point $x^*$ via the goal inference step, and then the system drives the state towards that set point. Within the convex hull, users cannot control the robot directly; rather, they move this (invisible) set point, and then system responds. Controlling this system is like driving a car and controlling its speed only through cruise control. The layer of indirection may also explain why users remain unsatisfied with the system despite improvements in trial metrics~\cite{javdani18} and particularly dislike pure assistance without any arbitration with the user's raw input~\cite{newman2022harmonic}.
This analysis also explains the equilibrium that users discovered intuitively in \citet{Aronson2024IntentionalAssistance}: by carefully altering the set point, users can cause the system to stop at any given location within the convex hull of the goals.

\section{Discussion and Conclusion}
In this paper, we perform a dynamical systems analysis of the shared autonomy framework. Using this analysis technique, we identify a limitation of the system when given a constant input and propose alternate formulations to fix the behavior. We show that the assistance inherently restricts the user from achieving all possible actions available to the system. In a free-space navigation setting, which corresponds to the scenario in many user studies, we find that the assistance introduces a level of indirection between the user and controls, which enables intuitive identification of equilibria but may decrease user opinion.

The analysis here is an example of what this technique may provide to the field of shared control, particularly for goal-predictive assistance. Since the user experiences the robot with or without assistance as a black-box input-output system, the dynamical analysis can be applied to many other types of shared control. In general, this analysis transforms shared control algorithm design into the inverse of a controls problem: given that somebody will be controlling over a system, design the system dynamics that best enables that operator to achieve their goals. We look forward to expanding this type of analysis to other shared control algorithms and using it to develop design principles for general assistive systems.

This analytical technique is not a substitute for other ways of designing shared control algorithms. Instead, its role is to enable an analysis of the assistance system outside of its assumptions. While assumptions about user behavior are the most clearly affected by user adaptation, this technique enables examining the user's experience in the case that the user is performing a different task than the assistance is modeling or has a conflicting environment model with the assistance. It even provides a framework for considering ``adversarial'' user behavior from a neutral viewpoint to determine how well the assistance performs outside its intended user behavior. Future work involves using this technique to develop general analysis tools for shared control systems that apply independently of many assumptions.

This paper presents a method for analyzing shared control systems and developing design principles for useful assistance. This user-centered, control-theoretic approach promises to enable systems that expand people's capabilities, the true goal of shared control.

\section*{Acknowledgment}
This material is based upon work supported by the National Science Foundation under grant no. IIS-2132887.

\bibliographystyle{IEEEtranN}
\bibliography{references}

% Generated by IEEEtranN.bst, version: 1.14 (2015/08/26)
\begin{thebibliography}{27}
\providecommand{\natexlab}[1]{#1}
\providecommand{\url}[1]{#1}
\csname url@samestyle\endcsname
\providecommand{\newblock}{\relax}
\providecommand{\bibinfo}[2]{#2}
\providecommand{\BIBentrySTDinterwordspacing}{\spaceskip=0pt\relax}
\providecommand{\BIBentryALTinterwordstretchfactor}{4}
\providecommand{\BIBentryALTinterwordspacing}{\spaceskip=\fontdimen2\font plus
\BIBentryALTinterwordstretchfactor\fontdimen3\font minus
  \fontdimen4\font\relax}
\providecommand{\BIBforeignlanguage}[2]{{%
\expandafter\ifx\csname l@#1\endcsname\relax
\typeout{** WARNING: IEEEtranN.bst: No hyphenation pattern has been}%
\typeout{** loaded for the language `#1'. Using the pattern for}%
\typeout{** the default language instead.}%
\else
\language=\csname l@#1\endcsname
\fi
#2}}
\providecommand{\BIBdecl}{\relax}
\BIBdecl

\bibitem[Aronson and Short(2024)]{Aronson2024IntentionalAssistance}
\BIBentryALTinterwordspacing
R.~M. Aronson and E.~S. Short, ``{Intentional User Adaptation to Shared Control
  Assistance},'' \emph{ACM/IEEE International Conference on Human-Robot
  Interaction}, pp. 4--12, 3 2024. [Online]. Available:
  \url{https://dl.acm.org/doi/10.1145/3610977.3634953}
\BIBentrySTDinterwordspacing

\bibitem[Losey et~al.(2018)Losey, McDonald, Battaglia, and
  O’Malley]{Losey2018AInteraction}
\BIBentryALTinterwordspacing
D.~P. Losey, C.~G. McDonald, E.~Battaglia, and M.~K. O’Malley, ``{A review of
  intent detection, arbitration, and communication aspects of shared control
  for physical human–robot interaction},'' 1 2018. [Online]. Available:
  \url{http://asmedigitalcollection.asme.org/appliedmechanicsreviews/article-pdf/70/1/010804/5964415/amr_070_01_010804.pdf}
\BIBentrySTDinterwordspacing

\bibitem[Javdani et~al.(2018)Javdani, Admoni, Pellegrinelli, Srinivasa, and
  Bagnell]{javdani18}
\BIBentryALTinterwordspacing
S.~Javdani, H.~Admoni, S.~Pellegrinelli, S.~S. Srinivasa, and J.~A. Bagnell,
  ``{Shared autonomy via hindsight optimization for teleoperation and
  teaming},'' \emph{The International Journal of Robotics Research}, vol.~37,
  no.~7, pp. 717--742, 6 2018. [Online]. Available:
  \url{https://doi.org/10.1177/0278364918776060}
\BIBentrySTDinterwordspacing

\bibitem[Cimolino and Graham(2022)]{Cimolino2022TwoControl}
\BIBentryALTinterwordspacing
G.~Cimolino and T.~C. Graham, ``{Two Heads Are Better Than One: A Dimension
  Space for Unifying Human and Artificial Intelligence in Shared Control},'' in
  \emph{Conference on Human Factors in Computing Systems - Proceedings}.\hskip
  1em plus 0.5em minus 0.4em\relax Association for Computing Machinery, 4 2022.
  [Online]. Available: \url{https://dl.acm.org/doi/10.1145/3491102.3517610}
\BIBentrySTDinterwordspacing

\bibitem[Muelling et~al.(2017)Muelling, Venkatraman, Valois, Downey, Weiss,
  Javdani, Hebert, Schwartz, Collinger, and Bagnell]{muelling2017}
\BIBentryALTinterwordspacing
K.~Muelling, A.~Venkatraman, J.-S.~S. Valois, J.~E. Downey, J.~Weiss,
  S.~Javdani, M.~Hebert, A.~B. Schwartz, J.~L. Collinger, and J.~A. Bagnell,
  ``{Autonomy infused teleoperation with application to brain computer
  interface controlled manipulation},'' \emph{Autonomous Robots}, vol.~41,
  no.~6, pp. 1401--1422, 8 2017. [Online]. Available:
  \url{https://doi.org/10.1007/s10514-017-9622-4}
\BIBentrySTDinterwordspacing

\bibitem[Gopinath et~al.(2021)Gopinath, Javaremi, and
  Argall]{Gopinath2021CustomizedRobots}
\BIBentryALTinterwordspacing
D.~Gopinath, M.~N. Javaremi, and B.~Argall, ``{Customized Handling of
  Unintended Interface Operation In Assistive Robots},'' in \emph{Proceedings -
  IEEE International Conference on Robotics and Automation}, vol. 2021-May,
  2021, pp. 10\,406--10\,412. [Online]. Available:
  \url{https://github.com/argallab/interface_assessments
  http://arxiv.org/abs/2007.02092}
\BIBentrySTDinterwordspacing

\bibitem[Okamura(2004)]{Okamura2004MethodsSurgery}
\BIBentryALTinterwordspacing
A.~M. Okamura, ``{Methods for haptic feedback in teleoperated robot-assisted
  surgery},'' \emph{Industrial Robot}, vol.~31, no.~6, pp. 499--508, 2004.
  [Online]. Available: \url{https://pubmed.ncbi.nlm.nih.gov/16429611/}
\BIBentrySTDinterwordspacing

\bibitem[Jain and Argall(2019)]{Jain2019ProbabilisticRobotics}
\BIBentryALTinterwordspacing
S.~Jain and B.~Argall, ``{Probabilistic Human Intent Recognition for Shared
  Autonomy in Assistive Robotics},'' \emph{ACM Transactions on Human-Robot
  Interaction}, vol.~9, no.~1, pp. 1--23, 12 2019. [Online]. Available:
  \url{http://dl.acm.org/citation.cfm?doid=3375676.3359614}
\BIBentrySTDinterwordspacing

\bibitem[Baker et~al.(2007)Baker, Tenenbaum, and Saxe]{Baker2007GoalPlanning}
C.~L. Baker, J.~B. Tenenbaum, and R.~R. Saxe, ``{Goal Inference as Inverse
  Planning},'' in \emph{Proceedings of the Annual Meeting of the Cognitive
  Science Society}, no.~29, 2007, p.~29.

\bibitem[Ziebart et~al.(2008)Ziebart, Maas, Bagnell, and
  Dey]{Ziebart2008MaximumLearning}
B.~D. Ziebart, A.~Maas, J.~A. Bagnell, and A.~K. Dey, ``{Maximum Entropy
  Inverse Reinforcement Learning},'' in \emph{Proceedings of the 23rd National
  Conference on Artificial Intelligence - Volume 3}, ser. AAAI'08.\hskip 1em
  plus 0.5em minus 0.4em\relax AAAI Press, 2008, p. 1433–1438.

\bibitem[H{\"{u}}ntemann et~al.(2008)H{\"{u}}ntemann, Demeester, Nuttin, and
  Van~Brussel]{Huntemann2008OnlineSteering}
A.~H{\"{u}}ntemann, E.~Demeester, M.~Nuttin, and H.~Van~Brussel, ``{Online user
  modeling with Gaussian Processes for Bayesian plan recognition during
  power-wheelchair steering},'' in \emph{2008 IEEE/RSJ International Conference
  on Intelligent Robots and Systems, IROS}, 2008, pp. 285--292.

\bibitem[Dragan and Srinivasa(2013)]{Dragan2013AControl}
\BIBentryALTinterwordspacing
A.~D. Dragan and S.~S. Srinivasa, ``{A policy-blending formalism for shared
  control},'' in \emph{International Journal of Robotics Research}, vol.~32,
  no.~7.\hskip 1em plus 0.5em minus 0.4em\relax SAGE PublicationsSage UK:
  London, England, 6 2013, pp. 790--805. [Online]. Available:
  \url{http://journals.sagepub.com/doi/10.1177/0278364913490324}
\BIBentrySTDinterwordspacing

\bibitem[Reddy et~al.(2018)Reddy, Dragan, and Levine]{reddy2018rss}
\BIBentryALTinterwordspacing
S.~Reddy, A.~D. Dragan, and S.~Levine, ``{Shared Autonomy via Deep
  Reinforcement Learning},'' in \emph{Robotics: Science and Systems}, 2018.
  [Online]. Available: \url{http://www.roboticsproceedings.org/rss14/p05.pdf}
\BIBentrySTDinterwordspacing

\bibitem[Jun~Jeon et~al.(2020)Jun~Jeon, Losey, and
  Sadigh]{JunJeon2020SharedActions}
\BIBentryALTinterwordspacing
H.~Jun~Jeon, D.~Losey, and D.~Sadigh, ``{Shared Autonomy with Learned Latent
  Actions},'' in \emph{Robotics: Science and Systems}.\hskip 1em plus 0.5em
  minus 0.4em\relax Robotics: Science and Systems Foundation, 6 2020. [Online].
  Available: \url{https://doi.org/10.15607/rss.2020.xvi.011}
\BIBentrySTDinterwordspacing

\bibitem[Aronson and Admoni(2022)]{Aronson2022GazeTeleoperation}
R.~M. Aronson and H.~Admoni, ``{Gaze Complements Control Input for Goal
  Prediction During Assisted Teleoperation},'' in \emph{Robotics: Science and
  Systems}, 2022.

\bibitem[Marayong et~al.(2003)Marayong, Li, Okamura, and
  Hager]{Marayong2003SpatialFixtures}
P.~Marayong, M.~Li, A.~M. Okamura, and G.~D. Hager, ``{Spatial motion
  constraints: Theory and demonstrations for robot guidance using virtual
  fixtures},'' in \emph{Proceedings - IEEE International Conference on Robotics
  and Automation}, vol.~2, 2003, pp. 1954--1959.

\bibitem[Broad et~al.(2017)Broad, Murphey, and
  Argall]{Broad2017LearningDynamics}
\BIBentryALTinterwordspacing
A.~Broad, T.~Murphey, and B.~Argall, ``{Learning models for shared control of
  human-machine systems with unknown dynamics},'' in \emph{Robotics: Science
  and Systems}, vol.~13, 2017. [Online]. Available:
  \url{https://storage.googleapis.com/rss2017-papers/50.pdf}
\BIBentrySTDinterwordspacing

\bibitem[Yoneda et~al.(2023)Yoneda, Sun, Yang, Stadie, and
  Walter]{Yoneda2023ToAutonomy}
\BIBentryALTinterwordspacing
T.~Yoneda, L.~Sun, G.~Yang, B.~Stadie, and M.~R. Walter, ``{To the Noise and
  Back: Diffusion for Shared Autonomy},'' \emph{Robotics: Science and Systems},
  2023. [Online]. Available:
  \url{https://diffusion-for-shared-autonomy.github.io.}
\BIBentrySTDinterwordspacing

\bibitem[Rea and Seo(2022)]{Rea2022StillInterfaces}
D.~J. Rea and S.~H. Seo, ``{Still Not Solved: A Call for Renewed Focus on
  User-Centered Teleoperation Interfaces},'' \emph{Frontiers in Robotics and
  AI}, vol.~9, 2022.

\bibitem[Nikolaidis et~al.(2017)Nikolaidis, Hsu, and
  Srinivasa]{Nikolaidis2017Human-robotExperiments}
\BIBentryALTinterwordspacing
S.~Nikolaidis, D.~Hsu, and S.~Srinivasa, ``{Human-robot mutual adaptation in
  collaborative tasks: Models and experiments},'' \emph{International Journal
  of Robotics Research}, vol.~36, no. 5-7, pp. 618--634, 6 2017. [Online].
  Available:
  \url{https://journals.sagepub.com/doi/full/10.1177/0278364917690593}
\BIBentrySTDinterwordspacing

\bibitem[Parekh et~al.(2022)Parekh, Habibian, and Losey]{Parekh2022RILI:Intent}
\BIBentryALTinterwordspacing
S.~Parekh, S.~Habibian, and D.~P. Losey, ``{RILI: Robustly Influencing Latent
  Intent},'' in \emph{IEEE International Conference on Intelligent Robots and
  Systems}, vol. 2022-Octob, 2022, pp. 2135--2142. [Online]. Available:
  \url{https://youtu.be/lYsWM8An18g}
\BIBentrySTDinterwordspacing

\bibitem[Gopinath and Argall(2017)]{Gopinath2017ModeDisambiguation}
\BIBentryALTinterwordspacing
D.~E. Gopinath and B.~D. Argall, ``{Mode switch assistance to maximize human
  intent disambiguation},'' in \emph{Robotics: Science and Systems}, vol.~13,
  2017. [Online]. Available:
  \url{http://www.roboticsproceedings.org/rss13/p46.pdf}
\BIBentrySTDinterwordspacing

\bibitem[Jonnavittula and Losey(2022)]{Jonnavittula2022CommunicatingAutonomy}
\BIBentryALTinterwordspacing
A.~Jonnavittula and D.~P. Losey, ``{Communicating Robot Conventions through
  Shared Autonomy},'' in \emph{Proceedings - IEEE International Conference on
  Robotics and Automation}, 2 2022, pp. 7423--7429. [Online]. Available:
  \url{http://arxiv.org/abs/2202.11140}
\BIBentrySTDinterwordspacing

\bibitem[Zurek et~al.(2021)Zurek, Bobu, Brown, and
  Dragan]{Zurek2021SituationalAutonomy}
M.~Zurek, A.~Bobu, D.~S. Brown, and A.~D. Dragan, ``{Situational Confidence
  Assistance for Lifelong Shared Autonomy},'' in \emph{2021 IEEE International
  Conference on Robotics and Automation (ICRA)}, 2021, pp. 2783--2789.

\bibitem[Newman et~al.(2024)Newman, Paxton, Kitani, and
  Admoni]{Newman2024BootstrappingCollaboration}
\BIBentryALTinterwordspacing
B.~A. Newman, C.~Paxton, K.~Kitani, and H.~Admoni, ``{Bootstrapping Linear
  Models for Fast Online Adaptation in Human-Agent Collaboration},'' in
  \emph{AAMAS '24: Proceedings of the 23rd International Conference on
  Autonomous Agents and Multiagent Systems}, 2024. [Online]. Available:
  \url{https://dl.acm.org/doi/10.5555/3635637.3663006}
\BIBentrySTDinterwordspacing

\bibitem[Javdani et~al.(2015)Javdani, Srinivasa, and
  Bagnell]{Javdani2015SharedOptimization}
\BIBentryALTinterwordspacing
S.~Javdani, S.~Srinivasa, and A.~Bagnell, ``{Shared Autonomy via Hindsight
  Optimization},'' in \emph{Robotics: Science and Systems XI}, vol.~11.\hskip
  1em plus 0.5em minus 0.4em\relax Robotics: Science and Systems Foundation, 7
  2015. [Online]. Available:
  \url{http://www.roboticsproceedings.org/rss11/p32.pdf}
\BIBentrySTDinterwordspacing

\bibitem[Newman et~al.(2022)Newman, Aronson, Srinivasa, Kitani, and
  Admoni]{newman2022harmonic}
\BIBentryALTinterwordspacing
B.~A. Newman, R.~M. Aronson, S.~S. Srinivasa, K.~Kitani, and H.~Admoni,
  ``{HARMONIC: A Multimodal Dataset of Assistive Human-Robot Collaboration},''
  \emph{The International Journal of Robotics Research}, vol.~41, no.~1, pp.
  3--11, 7 2022. [Online]. Available:
  \url{http://journals.sagepub.com/doi/10.1177/02783649211050677}
\BIBentrySTDinterwordspacing

\end{thebibliography}
\balance

\end{document}